\documentclass[10pt,twocolumn,letterpaper]{article}

\usepackage{wacv}
\usepackage{times}
\usepackage{epsfig}
\usepackage{graphicx}
\usepackage{amsmath}
\usepackage{amssymb}
\usepackage{multirow}
\usepackage{booktabs} 
\usepackage{pifont}
\usepackage{xcolor}
\usepackage{tabu}
\usepackage[accsupp]{axessibility}
\newcommand{\cmark}{\text{\ding{51}}}
\newcommand{\xmark}{\text{\ding{55}}}

\wacvfinalcopy 

\ifwacvfinal
\fi

\ifwacvfinal
\usepackage[breaklinks=true,bookmarks=false]{hyperref}
\else
\usepackage[pagebackref=true,breaklinks=true,colorlinks,bookmarks=false]{hyperref}
\fi

\def\I{\mathbf{I}}
\def\D{\mathbf{D}}
\def\A{\mathbf{A}}
\def\O{\mathbf{O}}

\def\M{\mathbf{M}}

\def\R{\mathbb{R}}

\def\ie{i.e\onedot}

\def\vs{vs\onedot}

\begin{document}

\title{Beyond Mono to Binaural: Generating Binaural Audio from Mono Audio with Depth and Cross Modal Attention}

\author{Kranti Kumar Parida\textsuperscript{1} \hspace{2em} Siddharth Srivastava\textsuperscript{2} \hspace{2em} Gaurav Sharma\textsuperscript{1,3} \vspace{0.5em} \\
\textsuperscript{1 }IIT Kanpur \hspace{1.5em}
\textsuperscript{2 }CDAC Noida \hspace{1.5em}
\textsuperscript{3 }TensorTour Inc. \vspace{0.2em} \\
{\small \texttt{\{kranti, grv\}@cse.iitk.ac.in, siddharthsrivastava@cdac.in}}}

\maketitle

\begin{abstract}
   Binaural audio gives the listener an immersive experience and can enhance augmented and virtual reality. However, recording binaural audio requires specialized setup with a dummy human head having microphones in left and right ears. Such a recording setup is difficult to build and setup, therefore mono audio has become the preferred choice in common devices. To obtain the same impact as binaural audio, recent efforts have been directed towards lifting mono audio to binaural audio conditioned on the visual input from the scene. Such approaches have not used an important cue for the task: the distance of different sound producing objects from the microphones. In this work, we argue that depth map of the scene can act as a proxy for inducing distance information of different objects in the scene, for the task of audio binauralization. We propose a novel encoder-decoder architecture with a hierarchical attention mechanism to encode image, depth and audio feature jointly. We design the network on top of state-of-the-art transformer networks for image and depth representation. We show empirically that the proposed method outperforms state-of-the-art methods comfortably for two challenging public datasets FAIR-Play and MUSIC-Stereo. We also demonstrate with qualitative results that the method is able to focus on the right information required for the task. The project details are available at \url{https://krantiparida.github.io/projects/bmonobinaural.html}
\end{abstract}

\section{Introduction}
\label{sec:intro}
Humans can infer approximate location of different objects by hearing the sound they emit. This is possible because of two ears and the separation between them. Due to this separation there is a difference in sound waves received by both ears in terms of amplitude and time. These differences, known as Interaural Level Difference (ILD) and Interaural Time Difference (ITD), are exploited by the brain to infer spatial properties eg.\ position of the sound source \cite{rayleigh1875our}. Thus, while audio recorded with a single microphone loses all such characteristics whereas \emph{binaural} audio recreates original sound more accurately and gives the listener a feeling of being in the recording place. 

The recording setup for binaural audio requires two microphones placed inside a dummy human head's ears. Such setup is closer to human hearing as it accurately models the sound reflection around the head and within the folds of the ear. Since binaural recording requires a full size dummy head, it is too bulky to be integrated into standard devices such as cameras or smartphones. However, we could get high quality binaural audio using standard handheld devices if we could lift mono audio to binaural audio. 

The aim of this work is to tackle this problem of audio binauralization: take a mono channel audio as input and predict the corresponding two channel binaural audio. In most prior approaches \cite{gao20192, lu2019self, zhou2020sep, xu2021visually}, the visual information in the form of RGB image is used, in addition to the mono audio, to predict the binaural audio. The RGB image serves as an important side information for encoding appearance of the sound producing sources and their relative locations in the scene. But most existing approaches ignore other important information, like the distance of the source from the microphone or the geometry of the scene. In \cite{richard2020neural, gebru2021implicit} similar information in the form of explicit position and orientation of both source and receiver were fed along with the audio input. This improved the performance of the system as compared to using RGB images only. However, doing so requires specialized equipment to track position of the source(s) as well as the listener, which is infeasible in general. We address this by using depth features of the scene along with the visual appearance features as auxiliary signals in the process of audio binauralization. Further, image, depth and binaural audio have also been shown to be interrelated in \cite{parida2021beyond}. However, unlike \cite{parida2021beyond}, where the authors used binaural echoes to improve the depth prediction, here we perform the reverse task of using the depth features to obtain binaural audio. Here, depth features can be considered as a proxy for encoding both position information of sources and geometry of the scene.

As opposed to the prior approaches \cite{gao20192, zhou2020sep}, we use visual transformer \cite{ranftl2021vision} instead of convolutional layers as the backbone for extracting both visual and depth features. We propose a carefully designed cross-modal attention network to better associate different audio components present in the sound to the location and depth or the corresponding objects in the scene. We also separate magnitude and phase losses for the predicted audio. We do this as both these losses are very different mathematical function and operate in different range and factorizing them makes the learning easier. We evaluate our approach on two challenging public datasets for the task, ie.\ FAIR-Play and MUSIC-Stereo. We show that our approach outperforms the previous state-of-the-art approaches quantitatively, and produces meaningful interpretable qualitative results. 

\section{Related Work}
\label{sec:related_work}
\textbf{Audio-Visual Binaurlization: }Recently the task of audio binauralization has been attempted in a data-drive fashion \cite{gao20192,lu2019self, morgadoNIPS18, zhou2020sep} as compared to earlier approaches that use signal processing techniques \cite{jianjun2015natural, savioja1999creating, zotkin2004rendering}. All the signal processing approaches model the system in the form of a Linear Time Invariant (LTI) system. In most of the cases \cite{jianjun2015natural, zotkin2004rendering}, HRTFs are measured and then convolutions are performed with them to get the final binaural audio.  More recently, data driven approaches have been tried for the task. In all the recent data driven approach some form of image information as auxiliary data have been used. In \cite{gao20192}, the authors have used RGB frame as side information for binaural audio generation. In \cite{lu2019self}, the authors have used both the RGB frame and optical flow along with audio features for binaural audio generation. In the similar lines, the authors in \cite{morgadoNIPS18}, have used both RGB and optical flow for generating full First Order Ambisonics for $360$-deg.\ videos. In \cite{zhou2020sep}, the authors approached the problem of audio binauralization in a multi-task setting by combining the task of source separation with it. In \cite{rachavarapu2021localize}, the authors have improved the task of audio binaurlization by performing localization of sound sources in the image. In \cite{gebru2021implicit, richard2020neural}, the authors performed binauralization of speech and noise signal played using a speaker by explicitly using the position and orientation of source and listener along with the audio features. A preliminary investigation of the usefulness of depth features for the task of binauralization is also available in \cite{parida2021depth}.

\textbf{Binaural audio and Depth: }There is an inherent interplay between binaural audio and the depth of the objects in the scene. As our aim in the paper is to improve the task of bianauralization using depth information, the reverse task, i.e.\ improving depth prediction from binaural audio has also been attempted. We give here some of the recent works in this line. In \cite{christensen2020batvision}, the depth map of the scene is estimated directly from received bianural echoes. In \cite{gao2020visualechoes, parida2021beyond}, authors have used the received binaural echoes along with images to improve upon the task of depth prediction from images. In similar lines, the authors in \cite{vasudevan2020semantic} have solved both the audio spatialization and depth prediction in multi-task framework. Here, instead of echoes, a two channel audio directly from the sound source is used for both depth estimation of the scene and audio spatialization, where the number of audio channels are increased to eight from two.

The depth of scene as an additional information has also been shown to be useful in other tasks such as image relighting \cite{liu2021relighting}, 3D pose estimation \cite{wu2019futurepose} etc.

\textbf{Audio-Visual Learning: } As our proposed work comes under the broad area of audio-visual processing, we give here some recent works in this area. In most of the works, semantic \cite{arandjelovic2017look, arandjelovic2018objects, morgado2021audio, parida2021discriminative}, temporal \cite{owens2018audio, korbar2018cooperative}, and spatial correspondence \cite{morgado2020learning, yang2020telling} between between both the modalities have been explored for learning features individually for each modality in a self-supervised manner. A separate stream of research have fused information from both audio and visual modality to improve upon tasks such as audio source separation \cite{ephrat2018looking, gan2020music, gao2018learning, zhao2018sound}, sounding object localization \cite{afouras2020self, hu2020discriminative, senocak2018learning}, zero-shot learning \cite{mazumder2020avgzslnet, parida2020coordinated}, saliency prediction \cite{tsiami2020stavis}. 
\section{Approach}
\label{sec:approach}
\begin{figure}
    \centering
    \includegraphics[scale=0.6]{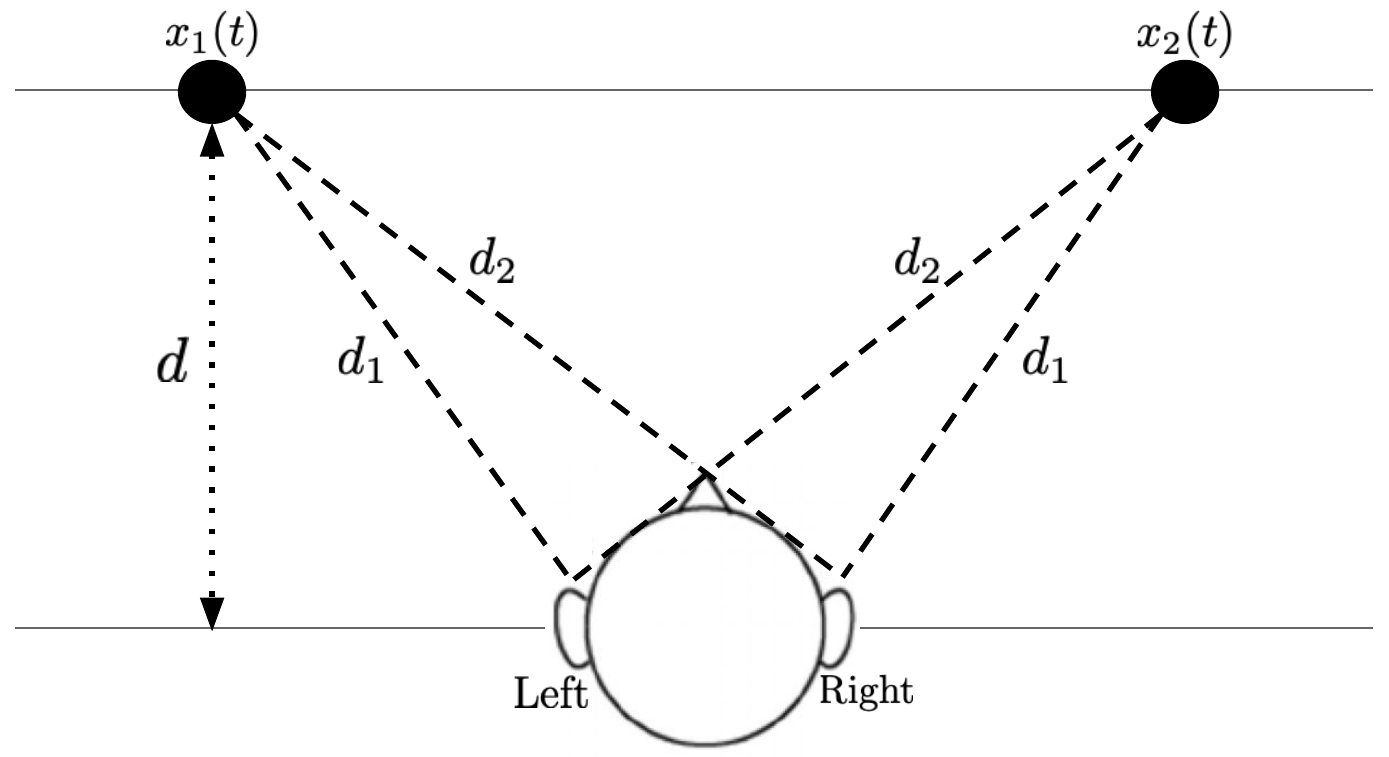}
    \caption{\textbf{Illustration of the concept.} $x_1(t)$ and $x_2(t)$ are two sound sources located at a distance $d$ from the recording device. The sound received by the left and right ears will be different because of the head shape and the depth of the sound producing source, both in amplitude and time axes. Human brain exploits these differences for inferring the spatial information of the sources.}
    \vspace{-1 em}
    \label{fig:illustration}
\end{figure}

\begin{figure*}
    \centering
    \includegraphics[scale=0.29]{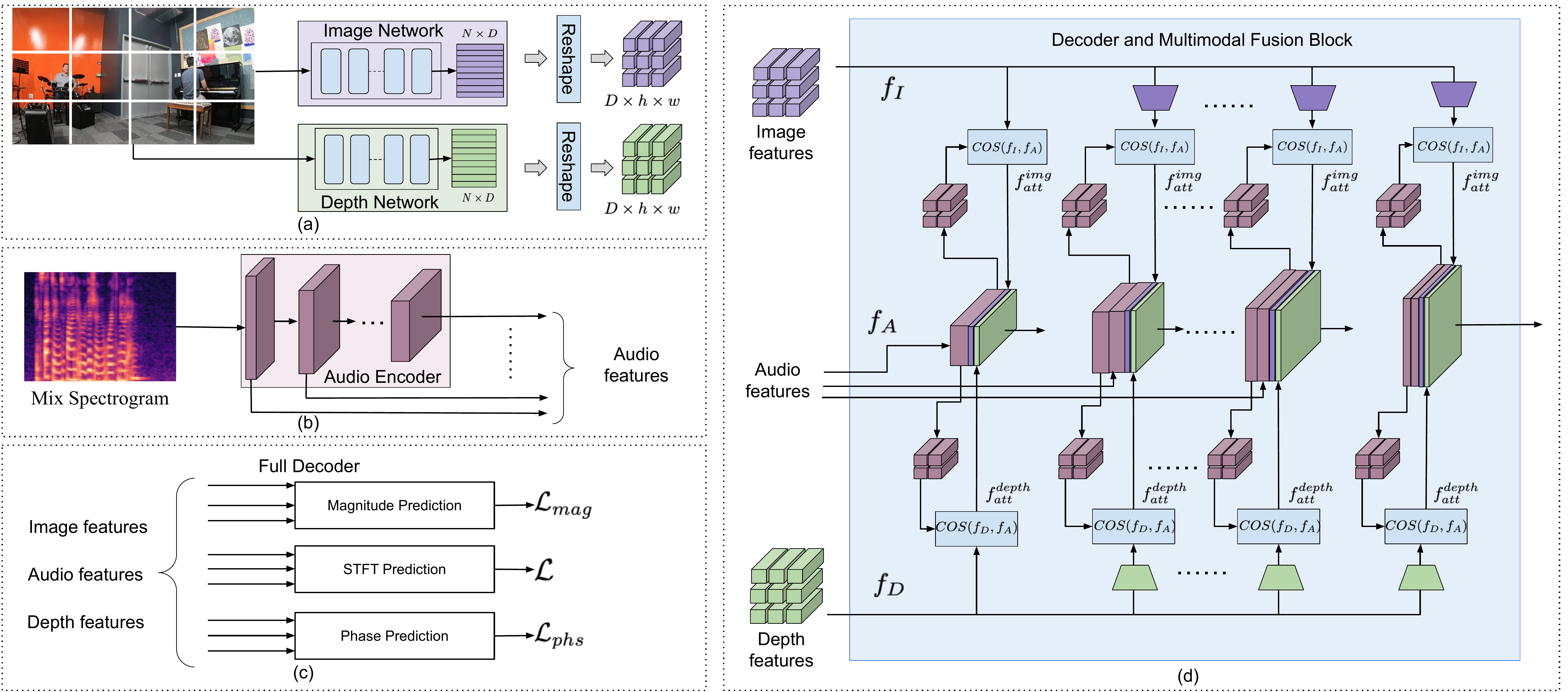}
    \caption{\textbf{Block diagram of proposed architecture.} The network takes mono audio and RGB image as input, and produces corresponding binaural audio consistent with the visual scene. (a) Image and Depth network input the same RGB image producing image and depth features,  $f_\I$ and $f_\D$ respectively. Similarly, (b) audio encoder inputs mono audio producing audio features $f_\A$. (c) The image, audio and depth features are fed into individual decoder blocks having individual subnetworks to predict magnitude, STFT and phase of the difference of both channels. Each decoder has the same architecture. The detailed architecture of one such decoder is shown in (d). In each decoder, cross modal attention is computed for both image-audio and depth-audio at each layer. Both the attention outputs are then concatenated with audio features to obtain the final predicted binaural audio. We also use skip connections to concatenate features from audio encoder layer to each layer of the decoder.} \vspace{-1 em}
    \label{fig:block_diag}
\end{figure*}

Our task is to convert a mono channel audio, $x(t)$ to a binaural audio with $(y_l(t), y_r(t))$ as the left and right channels respectively. To achieve this, we design a transformer network based deep neural network with three input modalities, ie., RGB image, depth and mono channel audio. Using this multimodal network, we exploit inherent relationship between the two channels of audio and the sound source's distance and relative location in the scene.

Consider a simple case of two sound sources in the scene $x_1(t)$ and $x_2(t)$, at a depth $d$, with one in the extreme left of visual field and other in the extreme right as in Fig.\ref{fig:illustration}. The mono audio, $x(t)$, is the combination of two sources, \ie $x(t) = x_1(t)+x_2(t)$. Now, when the sound is received at each ear, there will be a time difference between the arrival of $x_1(t)$ and $x_2(t)$. For the left ear, $x_1(t)$ will arrive earlier than $x_2(t)$ and the reverse for the right ear. Assuming no reflecting and absorbing materials in the scene, the direct sound received at left and right ear can be modelled as
\begin{align}
    y_l(t) &= \alpha_1 x_1(t-t_1) + \alpha_2 x_2(t-t_2)\\
    y_r(t) &= \alpha_1 x_1(t-t_2) + \alpha_2 x_2(t-t_1),
\end{align}
where, $t_1$, $t_2$ are the time delays with $t_1 < t_2$, and $\alpha_1, \alpha_2$ are the amplitude scaling factors. The time delays are symmetric wrt both ears because of the symmetric placement of sound sources. Let the distance for the left and right ear be $d_1$ and $d_2$ for source $x_1$. Similarly, the distance for right sound source will be $d_2$ and $d_1$ for left and right ear. There is a direct relationship between $t_1$ and $d_1$, i.e.\ $t_i = \frac{d_i}{v_s} \forall i=1,2$, where $v_s$ is the velocity of sound in air. The amplitude scaling factor, $\alpha_1, \alpha_2$, also have a direct relationship with the distance as the wave attenuates more as it travels longer distance. The ITD and ILD described in the previous section are due to $t_1, t_2$ and $\alpha_1, \alpha_2$ respectively.

Hence, for the network to predict realistic binaural audio it should effectively model ITD and ILD. This depends upon the relative arrangement of sound sources and its distance from the recording device. Taking note of this fact, we use both depth and image features of the underlying visual scene to infuse depth and position information of different sound sources in the prediction process. To achieve this, we propose a network consisting of carefully designed cross-modal attention mechanism to associate features from RGB, mono channel audio and depth. 

Following prior works \cite{gao20192, zhou2020sep, xu2021visually}, we use mix of both channels, $x_m(t) = x_l(t)+x_r(t)$ as input. The mixing of both channels looses spatial properties, and hence is a mono audio signal. For the output instead of predicting the individual left and right channels, we predict the difference between them, ie.\ $x_o(t) = x_l(t)-x_r(t)$. Finally, we perform simple arithmetic manipulation to get back the individual signals, where, $\hat{x}_l = \frac{x_m+\hat{x}_o}{2}$ and $\hat{x}_r = \frac{x_m-\hat{x}_o}{2}$. We use data in frequency domain by performing STFT on the time domain signal. We represent the STFT of input as $\A = \mathcal{F}(x_m) \in \R^{2 \times F \times T}$ and STFT of output as $\O = \mathcal{F}(x_o) \in \R^{2 \times F \times T}$. Further, we obtain the magnitude ($\O_{mag}$) and phase ($\O_{phs}$) of the complex output signal, where $\O_{mag} = \lVert \O \rVert_2$ and $\O_{phs} = \tan^{-1}(\frac{Re(\O)}{Im(\O)})$.

\subsection{Overall Architecture}
We show the overall architecture of our approach in Fig.~\ref{fig:block_diag}. The network consists of (i) audio encoder network, (ii) image network, (iii) depth network and (iv) audio decoder network. The audio encoder network is a convolutional network that takes the mono audio as input and gives audio features as output. Both the image and depth network are self-attention transformer networks. Both the networks have the same architecture and take RGB image as input to produce image and depth features respectively. The audio decoder network further has three different sub network, where one outputs directly the complex STFT of the difference of both channels and other two predicts the magnitude and phase of the difference independently. We perform cross-modal attention both for image-audio and depth-audio features separately at each layer of all the subnetworks of audio decoder network. The output of the network is the difference between the right and left channel audios. This difference prediction impedes the networks tendency to copy the same audio to both the channels, as observed first by~\cite{gao20192}. We now describe each component below.

\subsection{Image and Depth Network}
For extracting image and depth features from RGB image, $\I \in \R^{3 \times H \times W}$, we use the recently proposed vision transformer (ViT) \cite{dosovitskiy2020image} backbone. We use ViT-Large architecture consisting of $24$ attention blocks for both image and depth features. Following \cite{ranftl2021vision}, we obtain features from four different layers of the transformer network, \ie $l \in \{6, 12, 18, 24\}$, to get information at the varying level of details. Finally to align the number of channels of depth, image and audio networks, we perform a $1 \times 1$ convolution on features from each of the four layers resulting in $d$ channels. We then concatenate features from all the four layers and obtain image and depth features $f_{\I} \in \R^{4d \times h \times w}$ and $f_{\D} \in \R^{4d \times h \times w}$ respectively, where $h \times w$ represents total number of patches in the image. We then use both the features as input for hierarchical attention calculation in the network. We initialize the image and depth network with ImageNet \cite{deng2009imagenet} and MIX6 \cite{ranftl2019towards} respectively.

\subsection{Audio Encoder Network}
Similar to \cite{gao20192}, our audio encoder network consists of a UNet \cite{ronneberger2015u} style convolutional encoder architecture. We convert the time domain audio signal $x_m(t)$ into a STFT representation and concatenate both the real and imaginary to be fed as input to audio network, \ie $\A \in \R^{2 \times F \times T}$, where $F$ and $T$ are the no.\ of time-frequency bins in STFT. We then pass it through successive layers of convolutions. Finally, we obtain audio features, $f_{\A} \in \R^{d_A \times f \times t}$ as the output of audio encoder.

\subsection{Audio Decoder and Multimodal Fusion}
\noindent\textbf{Audio Decoder.} We adapt and build upon the audio decoder proposed in \cite{gao20192}. The decoder is further divided into three subnetworks and all share the same architecture whereas the output is different for each of the network. Each of the subnetwork in the decoder consists of $5$ fractionally strided convolutional layers, which increases the dimension of input tensor successively at each layer. Each of the network takes the input from all three modalities. The first subnetwork, Mag subnetwork, predicts the magnitude of STFT of the difference signal, $\lvert \Tilde{\O} \rvert \in \mathbb R^{1 \times F \times T}$. The second network, STFT subnetwork, directly predicts the STFT of difference signal and produces a mask, $\M \in \mathbb R^{2 \times F \times T}$, with values in the range $[-1, 1]$. We obtain the final output to be the difference between right and left channel audio, $\Tilde{\O} \in \mathbb{R}^{2\times F\times T}$. We predict it by multiplying the mask, $\M$ with the mixed input signal, $\A$, \ie $\Tilde{\O} = \M \cdot \A$. The final subnetwork, Phs subnetwork, predicts the phase of the difference signal. Similar to STFT subnetwork, the Phs subnetwork predicts a mask with values in the range $[-1, 1]$ for each of the time-frequency bin in the spectrogram, \ie $\M_p \in \mathbb R^{1 \times F \times T}$. As the phase of any signal lie in the range $[-\pi, \pi]$, we multiply $M_p$ with $\pi$ to get the predicted phase, i.e. $\text{Phs}(\Tilde{\O}) = \pi \times M_p$. At the time of prediction, we use the STFT subnetwork for obtaining the final output.

\noindent\textbf{Multimodal Fusion.} We perform similar multimodal fusion for all three subnetworks. The fusion operation combines the information from all three inputs at different scales. We show the fusion approach in the right side of Fig.~\ref{fig:block_diag}. The image and depth features are already extracted from different layers. For audio, each time-frequency bin in the feature representation can be considered as an unique audio concept.Each of the audio concepts can act as basic building blocks of different audio sources present in the scene. These audio concepts represent different characteristics sound of the source. This characteristic sound can contain frequency variation within the source and a single audio concept can also contribute to multiple sound sources. E.g.\ the audio concept for a $7$ string \textit{guitar} can be the sound produced by each of the strings. Similarly an audio concept for \textit{acoustic guitar} can also be shared by \textit{electric or classical guitar} or by any similar sounding object such as \textit{piano} or \textit{saxophone}. So, our goal in multimodal fusion is to effectively associate  different audio concepts to different object regions. Similar to the time-frequency bin in the audio representation, each co-ordinate in the spatial domain of the visual/depth features corresponds to certain region in the image. If the region contains a sounding object then the corresponding audio component should be weighted and also the depth value in the region should be used for the final output. We calculate two attention maps (i) between the image and audio features, and (ii) between the depth and audio features for fusing the information.

We design the network such that the output channels of audio encoder network is equal to the output channels of image and depth network, i.e.\ $d_I = d_D = d_A = 4d$. The attention is calculated between every pair of points.
{
\small
\begin{align}
    f_{att}^{img}(i,j,k, l) &= \frac{f_\I(:,i,j)^Tf_\A(:,k,l)}{\sqrt{\lVert f_\I(:,i,j) \rVert_2^2}\sqrt{\lVert f_\A(:,k,l) \rVert_2^2}} \forall i,j,k,l \\
    f_{att}^{depth}(i,j,k, l) &= \frac{f_\D(:,i,j)^Tf_\A(:,k,l)}{\sqrt{\lVert f_\D(:,i,j) \rVert_2^2}\sqrt{\lVert f_\A(:,k,l) \rVert_2^2}} \forall i,j,k,l
\end{align}
}
where, $f_{att}^{img}, f_{att}^{depth} \in \R^{h \times w \times f \times t}$ are the image-audio and depth-audio attention respectively. We then resize the 4D attention tensors into a 3D tensors such that the resulting attention maps are of size $[(h \times w) \times f_i \times t_i]$, where $f_i, t_i$ are the input spatial feature dimension in the $i^{th}$ layer of decoder. For the first decoder layer $f_i, t_i$ is exactly equal to the dimension of the audio encoder output. After resizing operation the number of channel dimension in the feature map corresponds to all the distinct regions in the image. We interpret this as the attention weight in all the regions of the image for the particular audio concept. We obtain final attention map at $i^{th}$ decoder level by concatenating both of them over the spatial axis.
\begin{align}
    f_{att}^{i} = \mathrm{Concat}(\mathrm{Resize}(f_{att}^{img}, f_{att}^{depth}))
\end{align}

Next, we concatenate the attention map with the audio features and feed the result to the next layer of the decoder. We also use skip connection at each decoder layer and concatenate features from corresponding encoder layer except for the first layer of decoder. As audio network increases the feature dimension in each level, it also increases the time-frequency bins in the feature representation and hence the finer details in the audio comes successively with each decoding layer. In order to account for the coarse to fine representation of the audio we add the image and depth features similar to first layer to obtain the attention map. To perform the attention calculation at each layer, the feature channel of the image and depth should align with the channels of audio features. To perform the alignment we use a one-layer neural network followed by GELU non-linearity to make the feature dimensions of both the modalities equal. For matching the channels of audio and image feature, the one-layer neural network used for every layer has weights of dimension $[d_I, d_i]$ and $[d_D, d_i]$ for image and depth features respectively. Please note that for first layer of decoder, i.e.\ $i=1$, we do not use one-layer network as the channels are already aligned.

\subsection{Loss Function and Training}
We use three individual losses for each of the subnetworks in the decoder. For the STFT subnetwork, following earlier works~\cite{gao20192, zhou2020sep}, we use an L2 loss between the ground truth and network output, given as
\begin{equation}
    \mathcal{L}(\hat{\O}, \O) = \lVert \hat{\O} - \O \rVert_2^2
    \label{eq:stft}
\end{equation} 
where, $\hat{\O}, \O$ are ground truth and predicted difference between the left and right channel audio. Similar to the STFT subnetwork, we also minimize the L2 loss for magnitude and phase, given as
\begin{align}
    \mathcal{L}_{mag}(\hat{\O}_{mag}, \O_{mag}) &= \lVert \hat{\O}_{mag} - \O_{mag} \rVert_2^2 \\
    \mathcal{L}_{phs}(\hat{\O}_{phs}, \O_{phs}) &= \lVert \hat{\O}_{phs} - \O_{phs} \rVert_2^2
    \label{eq:mag_phs}
\end{align}
where, $\hat{\O}_{mag}, \hat{\O}_{phs}$ are the predicted magnitude and phase obtained from the respective network, ${\O}_{mag}, {\O}_{phs}$ are the magnitude and phase of the ground truth signal. It is to be noted here that minimizing STFT loss in eq.~\ref{eq:stft} also implicitly minimizes magnitude and phase. We have added individual magnitude and phase loss possibly penalizing each term twice as this was found to be helpful in prior work \cite{richard2020neural}. Further to enforce the reconstruction of magnitude and phase is correct, we add a reconstruction loss. Here, we reconstruct back the real and imaginary part of the spectrogram and force it to be closer to the ground truth. We calculate the reconstruction loss $\mathcal{L}_{rec}$ by estimating the real and imaginary part of the spectrogram using $\hat{\O}_{mag}\mathrm{Cos}(\hat{\O}_{phs})$ and $\hat{\O}_{mag}\mathrm{Sin}(\hat{\O}_{phs)}$ respectively. We obtain the reconstructed STFT, $\hat{\O}$ by concatenating both the real and imaginary channels. The reconstruction loss is the L2 loss between the reconstructed STFT, $\hat{\O}$ and original STFT $\O$. 
\begin{equation}
    \mathcal{L}_{rec} = \lVert \hat{\O}_{mag}e^{i\hat{\O}_{phs}} - \O \rVert_2^2
    \label{eq:reconsturction}
\end{equation}

The final loss function used for training is weighted combination of all the losses, given as
\begin{equation}
    \mathcal{L}_{tot} = \mathcal{L}+\alpha_{mag}\mathcal{L}_{mag}+\alpha_{phs} \mathcal{L}_{phs}+\alpha_{rec} \mathcal{L}_{rec}
\end{equation}
where, $\alpha_{mag}, \alpha_{phs}, \alpha_{rec}$ are the hyperparameter denoting weights of individual loss and are set empirically. We train the whole network in an end-to-end manner.
\section{Experiments}
\label{sec:experiments}
\noindent\textbf{Dataset.} We report results on two dataset FAIR-Play and MUSIC-Stereo. For the FAIR-Play dataset \cite{gao20192}, we use the five new splits as proposed in \cite{xu2021visually} for our experiments. The dataset of Music-Stereo was proposed in \cite{xu2021visually} by combining two existing datasets MUSIC-21 and MUSIC-duet proposed in \cite{zhao2018sound} originally for the task of source separation. As the youtube IDs for MUSIC-Stereo is not publicly available, we select the binaural videos only from MUSIC-21 and MUSIC-duet as mentioned in \cite{xu2021visually}. In order to select the binaural videos from both the dataset, we calculate sum of the difference of left and right channel audio and then set a threshold of $0.001$ for selecting videos with binaural audio. We considered those that have sum of difference more than $0.001$ between both channels as binaural and discarded the rest. We obtained $713$ unique videos to have binaural audio. We then divide the videos into 80-10-10 into train, validation and test. Following the setting of \cite{xu2021visually}, we split the videos into $10$ second clips and we obtain a total of $15026$ clips, around $8$x more than FAIR-Play. For input data representation and preprocessing, we follow the same settings as described in earlier works \cite{gao20192, zhou2020sep, xu2021visually}. 

\noindent\textbf{Metrics.} Following \cite{xu2021visually}, we use five different metric for evaluation. STFT distance measures the squared $L_2$ distance between short time Fourier transform (STFT) of each channels for ground truth and predicted audio. Envelope (ENV) distance measures the $L_2$ difference of the envelope of ground truth and predicted audio for both channels, where we calculate the envelope from time-domain audio signal using Hilbert transform. Similar to STFT distance, we obtain Magnitude (Mag) distance by calculating the squared $L_2$ distance between the magnitude of STFT for both channels of ground truth and predicted audio. For Phase (Phs), we measure the $L_1$ difference between the phase of ground truth and predicted difference signal. We also report the Signal-to-noise ratio (SNR) for the predicted binaural audio, where signal refers to the ground truth binaural audio and noise refers to the distance between ground truth and prediction.

\subsection{Ablation Study}
In this section, we give the ablation results of our approach to demonstrate following points. (i) Impact of depth features over the image features. (ii)Contribution of each term of the loss function on the performance. 

\noindent\textbf{Impact of adding depth.} To analyse the effectiveness of depth for the task of audio spatialization, we study the impact of  each modality on the performance of the network. We add each of the modality (image, depth) one by one to the network and then combine all the modalities to verify the contribution of each modality. For a fair comparison, we use exactly the same transformer architecture with equal number of parameters for both image and depth. The results are shown in Tab.~\ref{tab:ablation}. We report the performance in the \textit{split-1} of modified FAIR-PLAY dataset for all the models.
\begin{table}
    \centering
    \resizebox{\columnwidth}{!}{
    \begin{tabular}{c|c|c|c|c|c}
    \hline
    Modality & STFT ($\downarrow$) & ENV ($\downarrow$) & Mag ($\downarrow$) & Phs ($\downarrow$) & SNR ($\uparrow$) \\
    \hline \hline
    audio & 1.337 & 0.166 & 2.674 & 1.560 & 5.01 \\
    \hline
    +image & 1.332 & 0.161 & 2.665 & 1.499 & 5.102 \\
    +depth & 1.334 & 0.165 & 2.668 & 1.553 & 5.036 \\
    +image+depth& \textbf{1.158}& \textbf{0.155} & \textbf{2.316} & \textbf{1.487} & \textbf{5.670} \\
    \hline
    \end{tabular}
    }
    \caption{\textbf{Audio binauralization by combining different modalities.} Using audio only (audio), audio with image features (+image), audio with depth features (+depth) and combination of audio, image and depth features (+image+depth). $\downarrow$ and $\uparrow$ indicates lower is better and higher is better respectively. 
    }
    \label{tab:ablation}
    \vspace{-1 em}
\end{table}

From Tab.\ \ref{tab:ablation}, we observe that there is a improvement in performance for all the metric with image + mono audio as input over an audio only input (Tab.~\ref{tab:ablation}, row 1 vs row 2). The value of STFT, ENV, Mag, Phs decreases from $1.337, 0.166, 2.674, 1.560$ to $1.332, 0.161, 2.665, 1.499$ whereas the SNR value increases from $5.01$ to $5.102$. Similarly, we observe a decrease of STFT, ENV, Mag and Phs to $1.334, 0.165, 2.668, 1.553, 5.036$ from $1.337, 0.166, 2.674, 1.560$ and increase of SNR to $5.036$ from $5.01$ by using depth + mono audio as input as compared to audio input only (Tab \ref{tab:ablation}, row 1 vs row 3). This shows that both the image and depth features are helpful towards a better audio binauralization. As both the image and depth backbone contain exactly same number of parameters, the improvement in performance can be attributed to the information encoded in it. Adding image results in a better performance in all the metrics over mono audio only input as compared to depth. This could be owed to the presence of semantic information in the RGB images in the form of appearance and relative location of different sound producing regions. Although adding depth information alone doesn't perform as good as to the approach of adding image information only but it performs better than the approach of using mono audio only as input. This is possibly due to the fact that depth input has relative distance information within the scene, and results in better binauralization as compared to mono audio input. From this observation, we hypothesize that combining depth with RGB will provide more contextual information leading to better localization of sound sources by the network and in turn better performance in binauralization task. This is also evident with the empirical performance of adding both depth and image features along with audio, which results in a significant improvement in performance in all the metrics. There is an an improvement of $~13\%$ for STFT,  $\sim 6\%$ for ENV, $\sim 13\%$ for Mag, $\sim 5\%$ for phs and $\sim 13\%$ for SNR over mono audio input (Tab \ref{tab:ablation}, row 1 vs row 4). This observation confirms that both image and depth information are helpful for the task of binauralization.

\begin{table}
    \centering
    \resizebox{\columnwidth}{!}{
    \begin{tabular}{c c c c|c|c|c|c|c}
    \hline
    $\mathcal{L}$ & $\mathcal{L}_{mag}$ & $\mathcal{L}_{phs}$ & $\mathcal{L}_{rec}$ & STFT ($\downarrow$) & ENV ($\downarrow$) & Mag ($\downarrow$) & Phs ($\downarrow$) & SNR ($\uparrow$) \\
    \hline \hline
    \cmark & \xmark & \xmark & \xmark & 1.206 & 0.158 & 2.413 & 1.488 & 5.418 \\
    \cmark & \cmark & \cmark & \xmark & 1.185 & 0.157 & 2.401 & 1.481 & 5.497 \\
    \cmark & \xmark & \xmark & \cmark & 1.190 & 0.158 & 2.411 & 1.487 & 5.435 \\
    \cmark & \cmark & \cmark & \cmark & \textbf{1.171} & \textbf{0.156} & \textbf{2.342} & \textbf{1.478} & \textbf{5.573} \\
    \hline
    \end{tabular}
    }
    \caption{\textbf{Contribution of different losses on performance.} Performance after applying different combination of losses. We observe that adding all the losses gives the best performance. $\downarrow$ and $\uparrow$ indicates lower is better and higher is better respectively.
    }
    \label{tab:ablation_loss}
    \vspace{-1 em}
\end{table}
\begin{table*}
    \centering
 \resizebox{\textwidth}{!}{
    \begin{tabular}{lcccccccccc}
    \toprule
    & \multicolumn{5}{c}{FAIR-Play }& \multicolumn{5}{c}{MUSIC-Stereo} \\
    \cmidrule(lr){2-6} \cmidrule(lr){7-11}
    Method & $\text{STFT} \downarrow$ & $\text{ENV} \downarrow$ & $\text{Mag} \downarrow$ & $ \text{Phs} \downarrow$ & $\text{SNR} \uparrow $ & $\text{STFT} \downarrow$ & $\text{ENV} \downarrow$ & $\text{Mag} \downarrow$ & $\text{Phs} \downarrow$  & $\text{SNR} \uparrow$ \\
    
    \midrule
    Mono-Mono~\cite{xu2021visually} &1.024 &0.145 &2.049 &1.571 &4.968 &1.014 &0.144 &2.027 &1.568 &7.858 \\
    Mono2Binaural~\cite{gao20192, xu2021visually}  &0.917 &\textbf{0.137} &1.835 &1.504 &5.203 &0.942 &0.138 &1.885 &1.550 &8.255 \\
    
    PseudoBinaural (w/o sep.)~\cite{xu2021visually} &0.951 &0.140 &1.914 &1.539 & 5.037 &0.953 &0.139 &1.902 &1.564 &8.129 \\
    PseudoBinaural~\cite{xu2021visually} & 0.944 & 0.139 & 1.901 & 1.522 & 5.124 &0.943 &0.139 &1.886 &1.562 &8.198 \\
    \textbf{Ours} &\textbf{0.909} &0.139 &\textbf{1.819} &\textbf{1.479} &\textbf{6.397} & \textbf{0.670}& \textbf{0.108}& \textbf{1.340}& \textbf{1.538}& \textbf{10.754}\\
    \hline
    \hline
    \color{gray}
    Sep-Stereo~\cite{zhou2020sep, xu2021visually} & \color{gray}0.906 & \color{gray}0.136 & \color{gray} 1.811 & \color{gray} 1.495 & \color{gray} 5.221 & \color{gray}0.929 & \color{gray}0.135 & \color{gray}1.803 & \color{gray}1.544 & \color{gray}8.306 \\
     
    \color{gray} Augment-PseudoBinaural~\cite{xu2021visually} & \color{gray} 0.878 & \color{gray} 0.134 & \color{gray} 1.768 & \color{gray} 1.467 & \color{gray} 5.316 & \color{gray} 0.891 & \color{gray} 0.132 & \color{gray} 1.762 &\color{gray} 1.539 &\color{gray} 8.419 \\
    
    \bottomrule
    \end{tabular}
     }
    \caption{\textbf{Comparison with existing approaches} We report the results for existing approaches directly from \cite{zhou2020sep}. $\downarrow$ indicates lower is better and $\uparrow$ indicates higher is better. The method in the last two rows uses atleast $2$x more data than ours and solve both task of audio binauralization and source separation jointly. Our approach outperforms other approaches in similar setting in almost all the metric and also performs comparably to superior approaches mentioned in the last two rows of the table for FAIR-Play dataset and even outperforms for MUSIC-Stereo dataset. 
    }
    \label{tab:sota}
    \vspace{-1 em}
\end{table*}

\noindent\textbf{Contribution of different Losses.} In order to get the contribution of individual losses in the final performance of the network, we add various combinations of the loss. We report the performance in \textit{split-1} after $50$ epochs in Tab.~\ref{tab:ablation_loss}. From Tab.~\ref{tab:ablation_loss}, we observe that all the losses contribute equally to the performance. We observe that adding both magnitude and phase loss improves all the metrics, i.e.\ STFT, ENV, Mag, Phs, SNR from $1.026, 0.158, 2.413, 1.488, 5.418$ to $1.185, 0.157, 2.401, 1.481, 5.497$ respectively (row 1 vs row2 in Tab.~\ref{tab:ablation_loss}). This proves that minimizing explicit magnitude and phase loss improves the performance, also consistent with the observation reported in \cite{richard2020neural}. When adding only the reconstruction loss without magnitude and phase loss in comparison to adding explicit magnitude and phase loss performance drops to $1.190, 0.158, 2.411, 1.487, 5.435$ from $1.185, 0.157, 2.401, 1.481, 5.497$ for all the metrics i.e.\ STFT, ENV, Mag, Phs and SNR respectively. This observation shows that minimizing individual magnitude and phase loss is more relevant for the task as compared to using reconstruction loss. We conclude that as both the losses, i.e.\ magnitude and phase are calculated from very different mathematical function (magnitude is a quadratic function of real and imaginary value where as phase is a trigonometric function) and are also in very different value range, hence separating them into individual component helps in the training process. Finally adding all the losses gives the best performance in all the metrics with STFT, ENV, Mag, Phs and SNR of $1.171, 0.156, 2.342, 1.478$ and $5.573$ respectively.

\subsection{Comparison to state-of-the-art}
\noindent\textbf{Baseline and prior approaches.} We compare our approach against various baselines and competitive approaches on both the datasets. The baseline of Mono-Mono is a simple approach where the input audio is copied for both the left and right channel. The existing approach of Mono2Binaural \cite{gao20192} uses only the image features along with audio features as the input to the decoder and a simple concatenation method for fusing both the features. In another existing approach of sep-stereo \cite{zhou2020sep}, a multi-task approach of binaural prediction and source separation is trained jointly using a single backbone. The data used for training is also atleast $2$x more than the amount of data used for training our method or Mono2Binaural approach. Although this approach is not directly comparable to ours as it uses more data and also solves multiple tasks jointly, this can serve as an upper bound for us. In one of the recent self-supervised approach, PseudoBinaural \cite{xu2021visually} generated from mono audio were used for training instead of the recorded ones. There are a number of variants to this approach, in PseudoBinaural (w/o sep.) the generated audios were used for training the binaural prediction task only whereas in PseudoBinaural the generated binaural audio is used for training both binauralization and source separation like sep-stereo \cite{xu2021visually}. Finally, in Augment-PseudoBinaural both the real audios and the generated ones were used for training both binauralization and separation task jointly. This is the most superior method as it solves both tasks and also uses $2$x more data as compared to sep-stereo and $4$x more data than ours. Similar to sep-stereo this method is not directly comparable to ours, we report it here as an upperbound for our case. 

\noindent\textbf{Comparison on FAIR-Play dataset.} We report the results on FAIR-Play dataset for different existing and baseline approaches along with the proposed approach in Tab.\ref{tab:sota}. We report the results for all the methods by averaging over all the five splits in the modified FAIR-Play dataset. We report the results on baseline and all the prior approaches directly from \cite{xu2021visually}. We observe that our method outperforms all the existing methods that is trained for a single task and dataset similar to ours. We observe that our propose method obtains an performance improvement of $\sim11\%, \sim4\%, \sim11\%, \sim6\%$ and $\sim29\%$ for STFT, ENV, Mag, Phs and SNR respectively over the baseline of Mono-Mono. Also, our method outperforms the best performing method in similar setting, Mono2Binaural in four metrics out of five. We obtain STFT and magnitude value for our approach to be $0.909$ and $1.819$ respectively as compared to $0.917$ and $1.835$ for Mono2Binaural. We also obtain the value of $1.479$ and $6.397$ for Phs and SNR outperforming Mono2Binaural as well. We also observe that the performance of our approach is also in the similar ball park of approaches that solves multiple tasks and uses multiple datasets. The performance of our approach in the three metric of STFT, ENV, Mag is only worse by $\sim0.3\%, \sim2\%, \sim0.4\%$ for Sep-Stero \cite{zhou2020sep} where as we outperform it on the rest two metrics Phs and SNR by $\sim2\%$ and $\sim18\%$ respectively. Finally, for the Augment-Pseudo we are worse in four metric STFT, ENV, Mag by $\sim3\%$ and Phs by $\sim0.8\%$. But we outperform even this method on SNR metric by $\sim17\%$. This proves that our method is competitive enough for the task of binauralization as it outperforms superior approaches in some of the metrics.

\noindent\textbf{Comparison on MUSIC-Stereo dataset. } We also report the performance of our approach along with baseline and existing approach in Tab.~\ref{tab:sota}. Similar to the FAIR-Play dataset, all the results of prior approaches and baseline are reported directly from \cite{xu2021visually}. We observe that our method outperforms significantly both baseline and other approaches in similar setting. There is an improvement of $\sim34\%, \sim25\%, \sim34\%, \sim2\%$ and $\sim36\%$ in the metrics of STFT, ENV, Mag, Phs and SNR respectively over the baseline. The proposed approach also outperforms the best method in similar setting, i.e.\ Mono2Binaural by $\sim29\%, \sim22\%,  \sim29\%, \sim0.8\%$ and $\sim30\%$ for all the metrics STFT, ENV, Mag, Phs and SNR respectively. Further, we observe that our method also significantly outperforms the mulit-task and larger data approaches, i.e.\ sep-stereo and Augment-PseudoBinaural, in all the five metrics as well. The Music-Stereo dataset contains diverse in the wild music videos from Youtube whereas FAIR-Play dataset contains videos where all of them are recorded inside the same recording room with very minimal variation in the background. Our proposed approach outperforms existing method by a higher margin in MUSIC-Stereo dataset as compared FAIR-Play dataset, which suggests that our method generalizes well to unconstrained setting.

\begin{figure}
    \centering
    \includegraphics[width=0.99\columnwidth]{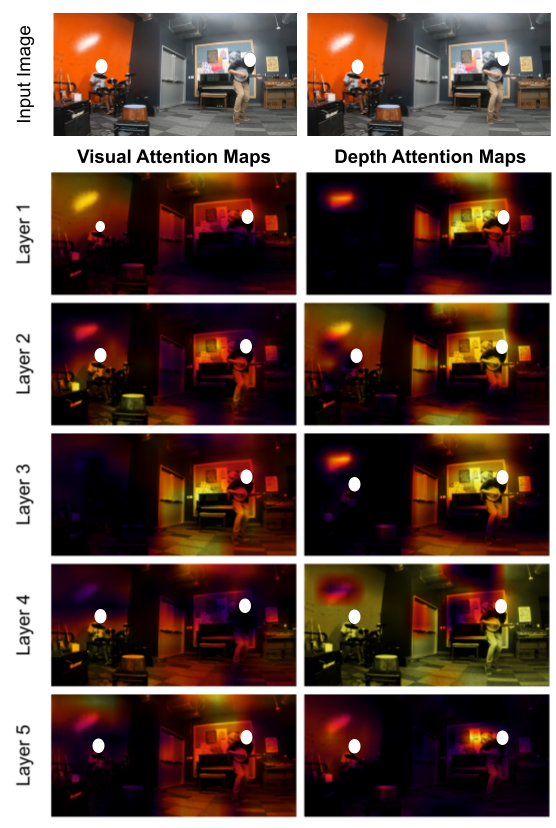}
    \caption{\textbf{Attention Map Visualization} Attention maps for both visual and depth channel at each decoder layer on FAIR-Play dataset. The first row shows the input image. We observe that visual attention map progressively attends to the sound producing regions in the image where as the depth attention maps attends to the structure of the room, i.e.\ wall, ceiling and floor.} 
    \label{fig:qual_viz}
    \vspace{-1em}
\end{figure}

\subsection{Qualitative Results} 
We give qualitative results of visual and depth attention map obtained from all layers of decoder in Fig.\ \ref{fig:qual_viz}. We provide the input image in the first row for comparison. From the visual attention map in first column, we observe that the attention values are spread out over the entire image in the first layer but in successive layers of 2,3, and 4 it produces high values only to the sound sources. We also observe that layer 3 produces high values for the source on the right side of the image whereas layer 4 produces high values for the source in the left side of the image. This region specific attention map can be considered as the inherent association between left and right audio channel with left and right regions of the image, which is important for an effective binauralization. For the depth attention maps, we observe that instead of attending to the sound source location, it looks at different structure of the rooms such as wall, ceiling and floor in layer 1, 2, and 3 of the decoder. From these attention maps, we make a general observation that the depth network infuses information about the geometry of the room resulting in better binauralization. We also provide predicted binaural results in our project page and request readers to listen to the videos to have a sense of the reconstruction.

\section{Conclusion}
We proposed an end-to-end trainable multi-modal transformer network with hierarchical multi-modal attention, for mono to binaural audio generation. We studied the impact of image and depth  inputs along with their combinations on this task. We demonstrated that adding depth provides additional structural information which significantly improves audio binauralization quantitatively and aids in better source localization qualitatively, as visually analysed from attention maps. The proposed method obtains state-of-the-art results on two challenging datasets (FAIR-Play and MUSIC-Stereo) for the task.

\vspace{0.1em} 
\noindent
\textbf{Acknowledgment.}  Kranti Kumar Parida gratefully acknowledges support from the Visvesvaraya PhD Fellowship.

\vspace{-0.5em}
{\small
\bibliographystyle{ieee_fullname}
\bibliography{egbib}
}

\clearpage
\appendix
\section{Ablation}
We provide further ablations for the outputs of the decoders. As mentioned in Fig.~2 of the main manuscript, our approach has three decoders each one predicting magnitude, STFT and phase of the difference of both channels individually. We use the STFT prediction branch for obtaining the final output binaural audio. We also experiment with a different architecture where there is a single decoder instead of three for predicting the output. We provide the results for different architectural variations in Tab.~\ref{tab:ablation_output}.

In Tab.~\ref{tab:ablation_output}, we give two different variants of the decoder architecture. In the first block of the table, we report the result for the case where there are three individual decoders. In the second block, instead of using three decoders we use a single decoder for obtaining the output. We note here that for both the cases we can add all the four different losses (Sec.~3.5 in main manuscript). But for the second case as we are not predicting magnitude and phase separately, we obtain them from the predicted STFT and use the same for magnitude, phase and reconstruction loss calculation. Hence, the output for second case is only the STFT prediction. We report the results for \textit{split-1} of modified FAIR-Play dataset after $50$ epochs.

We make two conclusions here. i) The addition of separate magnitude and phase prediction improves the performance (row-1 \vs row-4 in Tab.~\ref{tab:ablation_output}). We observe that there is significant improvement in performance of STFT, ENV from $1.301, 0.163$ to $1.171, 0.156$ with single decoder and three individual decoders respectively. We hypothesize that the primary reason for this are the very different mathematical functions for calculation of magnitude and phase as described in Sec~4.1 of main manuscript. ii) Adding individual Magnitude and Phase subnetworks regularizes the training process. As we are predicting magnitude and phase individually in the first case, we can also obtain the output by combining the predicted magnitude and phase from the respective networks. We give the results for both the cases in Tab.~\ref{tab:ablation_output} (row1 \vs row2). When we obtain the output from STFT branch, we get a performance of $1.171$, $0.156$ for STFT and ENV respectively, whereas combining predictions from Magnitude and Phase gives a performance of $1.267$ and $0.162$ respectively. This shows that using the predictions from STFT branch give better performance. However, using explicit magnitude and phase loss is beneficial as its removal degrades the performance to $1.206$ and $0.158$ (row3 in Tab.~\ref{tab:ablation_output}). This observation confirms that Magnitude and Phase subnetworks act as regularizers even though the prediction from these branches do not add directly over the STFT branch.

\begin{table}
    \centering
    \resizebox{\columnwidth}{!}{
    \begin{tabular}{l|c c c c|c|c}
    \hline
     Approach & 
    $\mathcal{L}$ & $\mathcal{L}_{mag}$ & $\mathcal{L}_{phs}$ & $\mathcal{L}_{rec}$ & STFT ($\downarrow$) & ENV ($\downarrow$) \\
    \hline \hline
    STFT & \cmark & \cmark & \cmark & \cmark & 1.171 & 0.156 \\
    Mag-Phs & \cmark & \cmark & \cmark & \cmark & 1.267 & 0.162 \\
    \hline
    STFT only &\cmark & \xmark & \xmark & \xmark & 1.206 & 0.158 \\
    STFT (w/ mag-phs) &\cmark & \cmark & \cmark & \cmark & 1.301 & 0.163 \\
    \hline
    \end{tabular}
    }
    \caption{\textbf{Comparison of different output setting.} The first block of results (rows 1,2) are for three individual decoders and the second block (rows 3,4) are for those using a single decoder only. We observe that adding all the losses with individual decoder and taking output directly from STFT gives the best performance. $\downarrow$ indicates lower value is better.
    }
    \label{tab:ablation_output}
    \vspace{-1 em}
\end{table}
\section{Qualitative Videos}
We give four qualitative videos obtained using our approach. We request the readers to look at the results available at our project page \url{https://krantiparida.github.io/projects/bmonobinaural.html}. We request the readers to use a high quality headphone and use both left and right speakers, to be able to appreciate the binaural audio. The first two files contain single sound producing source. In first video, the sound source is present towards the left and hence the audio is dominant in the left audio channel. In the second example, the sound is present towards the right side of the scene. Hence the audio is dominant in the right audio channel. We can also observe in this example that as the source move towards the centre, the predicted audio also follows the trajectory. This shows that our approach is able to model even the subtle variations in the input. In third example, there are two sources in the scene, one on the left and other on the right. Our network successfully produces the binaural audio where the audio for each of the source goes predominantly to the corresponding channels. We show a limitation of our approach in the final example, where the two sound sources produce similar sounds resulting in a relatively inferior binaural audio. However, the generated binaural audio is still perceptually better than the mono audio. Hence our approach can further be improved if we integrate some form of source separation knowledge/prior into the model, which we consider as a promising future direction.

\end{document}